%% file: main.tex
\documentclass[preprint,11pt,3p]{elsarticle}
\usepackage[utf8]{inputenc}

\usepackage{amssymb}
\usepackage{latexsym}
\usepackage{rotating, multirow}
\usepackage{makecell}
\usepackage{algorithm}
\usepackage{listings}
\usepackage{bm}
\usepackage{algorithmicx}
\usepackage[noend]{algpseudocode}
\usepackage{booktabs}
\usepackage{natbib}
\usepackage{graphicx}
\usepackage{subcaption}
\usepackage{amstext}
\usepackage{amsmath}
\usepackage{amssymb}
\usepackage{braket}
\usepackage{tikz}
\usepackage{svg}
\usepackage[super]{nth}
\usepackage{url}
\usepackage{xcolor}
\usepackage{hyperref}

\input{commands.tex}

\begin{document}
    \input{alt.tex}

    \input{front.tex}

    \input{content.tex}
    \input{acknowledgments.tex}
    \bibliographystyle{abbrvnat}\biboptions{authoryear}
    \bibliography{main}
\end{document}

%% file: commands.tex
\newcommand{\figref}[1]{Fig.~\ref{#1}}
\newcommand{\tabref}[1]{Tab.~\ref{#1}}
\newcommand{\equref}[1]{Eq.~\ref{#1}}
\newcommand{\secref}[1]{Sec.~\ref{#1}}
\newcommand{\algoref}[1]{Alg.~\ref{#1}}
\newcommand\br[1]{\left(#1\right)}
\newcommand\round[1]{\lfloor#1\rceil}
\newcommand\loss{\mathcal{L}}
\newcommand\abs[1]{\lvert #1 \rvert_1}

\newcommand{\fscore}[1]{F1\ensuremath{_{\tau={#1}}}}

\newcommand{\unet}{\textsc{U-Net}}
\newcommand{\mrcnn}[1]{\textsc{Mask R-CNN}-{#1}}
\newcommand{\maskrcnn}{\textsc{Mask R-CNN}}
\newcommand{\cpn}[2]{\textsc{CPN}\ensuremath{_{\mathcal{R}{#1}}}-{#2}}

\newcommand{\bbunet}{\textsc{U22}}
\newcommand{\bbr}[2]{\textsc{R}{#1}\textsc{-{#2}}}
\newcommand{\bbx}[2]{\textsc{X}{#1}\textsc{-{#2}}}
\newcommand{\bbbc}[1]{\textsc{BBBC0{#1}}}
\newcommand{\ncb}{\textsc{NCB}}
\newcommand{\synth}{\textsc{SYNTH}}

%% file: alt.tex
\makeatletter
\def\ps@pprintTitle{%
  \let\@oddhead\@empty
  \let\@evenhead\@empty
  \let\@oddfoot\@empty
  \let\@evenfoot\@oddfoot
}
\makeatother

%% file: front.tex
\begin{frontmatter}
    \title{Contour Proposal Networks for Biomedical Instance Segmentation}
    \author[fzj,hai]{Eric {Upschulte}\corref{cor1}}
    \cortext[cor1]{Corresponding author: Tel.: +49-2461-61-5960; fax: +49-2461-61-3483;}
    \ead{e.upschulte@fz-juelich.de}
    \author[hhu]{Stefan {Harmeling}}
    \author[fzj,br]{Katrin {Amunts}}
    \author[fzj,hai]{Timo {Dickscheid}}
    
    \address[fzj]{Institute of Neuroscience and Medicine (INM-1), Research Centre Jülich, Germany}
    \address[hai]{Helmholtz AI, Research Centre Jülich, Germany}
    \address[hhu]{Institute of Computer Science, Heinrich Heine University, Düsseldorf, Germany}
    \address[br]{C\'{e}cile \& Oscar Vogt Institute for Brain Research, University Hospital Düsseldorf, Germany}
    
    \begin{abstract}
        We present a conceptually simple framework for object instance segmentation called \emph{Contour Proposal Network} (CPN), which detects possibly overlapping objects in an image while simultaneously fitting closed object contours using an interpretable, fixed-sized representation based on Fourier Descriptors. The CPN can incorporate state of the art object detection architectures as backbone networks into a single-stage instance segmentation model that can be trained end-to-end. We construct CPN models with different backbone networks, and apply them to instance segmentation of cells in datasets from different modalities. In our experiments, we show CPNs that outperform U-Nets and Mask R-CNNs in instance segmentation accuracy, and present variants with execution times suitable for real-time applications. The trained models generalize well across different domains of cell types. Since the main assumption of the framework are closed object contours, it is applicable to a wide range of detection problems also outside the biomedical domain. An implementation of the model architecture in PyTorch is freely available.
    \end{abstract}
\end{frontmatter}

%% file: content.tex
\section{Introduction}
\label{introduction}

\begin{figure*}
    \centering
	\includegraphics[width=\linewidth]{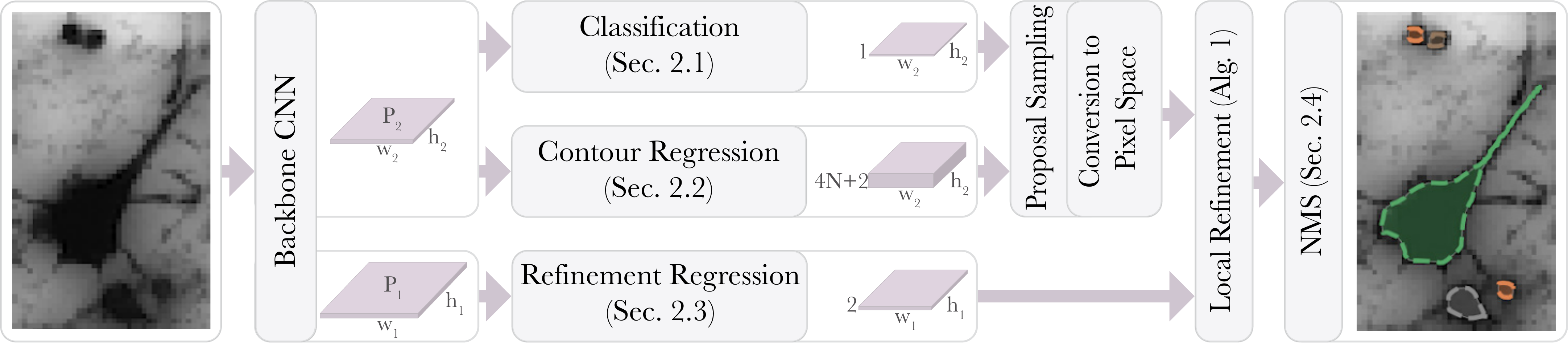}
	\caption{The Contour Proposal Network (\emph{CPN}) setup for instance segmentation. An initial backbone network computes feature maps $P_1$ and $P_2$. Based on the low-resolution $P_2$ a classification head determines for each pixel if an object is present or not, while the contour regression heads generate object contours, defined in the frequency domain, at each pixel. All contour representations that are classified to represent an object are extracted and converted to pixel space using \equref{eq:fourier}. The high-resolution $P_1$ is used to regress a refinement tensor that is used during a Local Refinement step (\algoref{alg:refine}) that maximizes pixel accuracy.
	Finally, non-maximum suppression (\emph{NMS}) is applied to remove redundant detections.}  
	\label{fig:model-mini}
\end{figure*}

\subsection{Motivation}

Instance segmentation is the task of labeling each pixel in an image with an index that represents distinct objects of predefined object classes. 
This is different from semantic segmentation, which assigns the object class itself to each pixel, and does not distinguish objects of the same type if their shapes touch or overlap. A common instance segmentation problem in biomedical imaging is the detection of cells in microscopic images, in particular for quantitative analysis.
While the pixel accuracy of recent cell segmentation methods has become sufficient for many imaging setups, detection accuracy often remains a bottleneck, especially wrt.~handling of touching and overlapping objects.
In many biomedical applications, accurate object detection and realistic recovery of object shape is both desirable. 
However, many instance segmentation methods define one unique object index per pixel, referring to the foreground object only. This results in an incomplete capture of partially superimposed objects, and consequently to a misrepresentation of their actual shape (as in e.g.~\figref{fig:systematic_errors}g top) which in turn might impair shape-sensible downstream tasks like morphological cell analysis.
To avoid such problems, instance segmentation methods with appropriate modeling of object boundaries are required. 

Furthermore, segmentation models should generalize well to variations in the data distribution. This is important for small variations, which inevitably occur in practical lab settings due to variations between different samples, fluctuations of histological protocols and digital scanning processes \citep{DBLP:journals/corr/abs-1909-11575, yagi_color_2011}.
Generalizability is also important at the scale of data domains, in order to allow transfer of trained models with manageable annotation efforts. 

\subsection{Related work}
\label{sec:related}

\paragraph{Pixel classifiers}
Instance segmentation can be achieved using a dense pixel classifier such as the U-Net \citep{DBLP:journals/corr/RonnebergerFB15}, and can be casted from a semantic segmentation solution to an instance-agnostic approach using a grouping strategy such as connected component labeling (CCL). This will group multiple pixels of the same class into non-overlapping instances. To distinguish touching instances as well, one may introduce narrow background gaps between objects with careful per pixel loss weightings \citep{DBLP:journals/corr/RonnebergerFB15}. Improved versions define border pixels as an additional class \citep{chen2016dcan, Guerrero_Pena_2018, Zabawa_2020}.
Such models have demonstrated to segment the borders of isolated objects very precisely. 
However, in case of crowded images, already a few falsely classified pixels can merge close-by instances and critically impair the detection result \citep{Caicedo335216}.

\paragraph{Pixel classifiers coupled with shape models}
To reach better robustness on crowded images, some authors proposed to couple active contour models with CNN-based segmentation models.
\cite{thierbach2018_CombiningDeepa} first employed a dense pixel classifier to predict probability maps of object centroids. These maps were then thresholded to initialize a subsequent active contour segmentation.
\cite{zhang2018_LearningDeep} suggested a scheme where a CNN is trained to explicitly predict the energy function for fitting an active contour model to a given object.
The contour computation is here attached as a black box to the learning loop, so that the conversion from pixel to shape space and back is invisible to the network training, and thus not part of the actual learning.
\cite{gur2019_EndEnd} proposed to use a neural renderer as a differentiable domain transition from polygon to pixels, allowing a full learning path. This way they train a U-Net-like CNN that produces 2D displacement fields for polygonal contour evolution with a loss that addresses the segmentation as well as ballooning and curvature minimizing forces in the pixel domain. 
While this allows end-to-end training, the actual boundary representation remains hidden and is not accessible to downstream tasks.

\paragraph{Dense vs sparse detectors}

The above-mentioned solutions have in common that they learn object masks in the pixel domain under a hidden or decoupled shape model, producing a dense classification by assigning a label to each pixel of an image. An alternative is to perform object detection by directly estimating the parameters of a contour model in its embedding space, and attaching a pixel location to the shape descriptor. This way the bounds of an entire object are concentrated at a single pixel, leading to a sparse detection scheme and forcing the model to develop an explicit internal understanding of instances. For closed contours, pixel masks can then be obtained by rasterization. Giving direct emphasis (and possibly supervision) to the shape model, such an approach could provide a more interpretable and  efficient problem representation.    

\paragraph{Bounding box regression}
The de-facto standard for modeling boundaries in object detection networks are bounding boxes \citep{ren2016faster, Liu_Anguelov_Erhan_Szegedy_Reed_Fu_Berg_2016, lin2018focal, redmon2018yolov3, bochkovskiy2020yolov4, yang2020_NuSeTDeep}.
Here, the models predict at least four outermost object locations. This approach captures little information about the object instance beyond location, scale and aspect ratio \citep{jetley2017straight}.
The most established approach from this category is the Mask R-CNN \citep{he2018mask}. It first detects bounding boxes by regression, and then gathers image features inside the bounding box to produce pixel masks.

\paragraph{Regression of shape representations}
More detailed shape representations have been proposed in recent years as well
\citep{jetley2017straight, schmidt2018_CellDetection, miksys2019straight, xie2020_PolarMaskSingle}.
Closest to our work is the approach of 
\cite{jetley2017straight}, who combined the popular YOLO architecture \citep{redmon2016look} with an additional regression of a decodable shape representation for each object proposal.
They showed that integration of a higher-order shape reasoning into the network architecture improves generalization. In particular, it allowed to predict plausible masks for previously unseen object classes.
They evaluated three different shape representations, namely fixed-sized binary shape masks, a radial representation, and a learned shape encoding. 
The binary shape masks show quantitatively worse results than the other two representations.
The radial representation defines a series of offsets between an anchor pixel and points on its contour, and turned out to be inferior for common object classes in natural images.
It has also been applied for cell nuclei detection in the \emph{StarDist} architecture \citep{schmidt2018_CellDetection}, which showed good detection accuracy but stays behind the pixel precision achieved by U-Nets \citep{DBLP:journals/corr/RonnebergerFB15}.
StarDist was extended as \emph{PolarMask} \citep{xie2020_PolarMaskSingle} to be applicable to multiclass problems, such as the COCO dataset \citep{lin2015microsoft}. Also it was coupled with a different loss and evaluated with multiple backbone architectures.
In general, the applicability of the radial model is limited to the "star domain", which excludes many non-convex shapes \citep{dietler_convolutional_2020}.
Predicting radial representations also involves a predefined number of rays, leading to possibly suboptimal sampling and limited precision of the contour \citep{schmidt2018_CellDetection}.
As a third representation, \citep{jetley2017straight} train an auto-encoder  on the Caltech-101 silhouettes dataset to learn a shape embedding for the detection network. 
\cite{miksys2019straight} extended this approach with an additional distance transform acting as a proxy between decoder and shape image, which allows to superimpose "discs" at every pixel location and hence mitigate the impact of falsely predicted pixels. They also considered the inclusion of the decoder in the training process, and showed quantitative improvements. 
However, this model still lacks optimal pixel precision.

\subsection{The Contour Proposal Network}
Based on existing strengths and weaknesses in the field, we here introduce the \emph{Contour Proposal Network} (CPN).
Similar in spirit to the approach of \cite{jetley2017straight}, 
it models instance segmentation as a sparse detection problem by performing regression of object shape representations at single pixel locations.
The model architecture is depicted in \figref{fig:model-mini}:
A backbone network derives feature maps from an input image. For each pixel of the feature map, regression heads generate a contour representation, while a classification head determines whether an object is present at a given location.
Based on the classifications, a proposal sampling stage then extracts a sparse list of contour representations.
By converting these to the pixel domain using the fully differentiable Fourier sine and cosine transformation, we implicitly enforce the contour representations to be defined in the frequency domain,
inspired by Elliptical Fourier Descriptors \citep{Kuhl1982EllipticFF}.
The resulting contour coordinates are optimized by a local refinement procedure to further maximize pixel precision using a residual field, produced from an additional regression head. This is similar in spirit to the displacements fields used by \cite{gur2019_EndEnd}, but integrates more naturally as the CPN already operates with near-final contour proposals in the pixel domain at this stage. 
The complete framework is trained end-to-end across all these stages.
As a final inference step, non maximum suppression removes redundant detections from the object proposals.

We train CPNs, that outperform U-Nets and Mask R-CNNs in instance segmentation accuracy in our experiments and demonstrate that inference speed of selected CPNs is suitable for real-time applications, especially when considering automatic mixed precision (\emph{amp}).
The trained models generalize well to other datasets which cover different families of biological cells.

\section{Methods}
\label{cpn}

The Contour Proposal Network (\textit{CPN}) uses five basic building blocks (\figref{fig:model-mini}).
Initially, dense feature maps $P_1\in \mathbb{R}^{w_1\times w_1 \times c_1}$ (high-resolution) and $P_2\in \mathbb{R}^{w_2\times w_2 \times c_2}$ (low-resolution) are generated by a \emph{backbone CNN} which can be freely chosen.
From the latent feature map $P_2$, a \emph{classifier head} detects objects, while parallel \emph{regression heads} jointly generate explicit contour representations. The classification scores generated by the classifier estimate whether an object exists at the given locations. Contours are modelled as a series of 2d coordinates by applying the Fourier sine and cosine transformation of degree $N$ to the latent outputs of the contour regression head, resulting in a fully differentiable, fixed-sized format for boundary regression (\secref{sec:contour_representation}). The contour proposals are a dense map on the pixel grid, with an $h_2\times w_2\times(4N+2)$ tensor of shape descriptors and an $h_2\times w_2 \times 1$ tensor of corresponding object classification scores. The output resolution $h_2 \times w_2$ is independent of the input resolution and effectively defines the maximum number objects that can be detected. All representations that are deemed to describe a present object are extracted as a list of contour proposals and mapped to pixel space using \equref{eq:fourier}. The proposals are then processed by a trainable \emph{refinement block} to maximize fit of contours with image content using high-resolution features $P_1$. The last building block filters redundant detections using non-maximum suppression.

\subsection{Detection} \label{sec:object_detection}
A classification head produces a detection score for each contour representation and states whether it represents a present object or not.
In our work we focus on the binary case and do not distinguish different object categories.

\subsection{Contour Representation} \label{sec:contour_representation}
\begin{figure}
	\centering
	\begin{subfigure}{0.32\textwidth}
		\includegraphics[width=\linewidth]{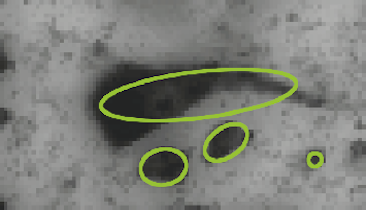}
		\caption{$N=1$, $6$d vector} \label{fig:ordera}
	\end{subfigure}
	\hspace*{\fill}
	\begin{subfigure}{0.32\textwidth}
		\includegraphics[width=\linewidth]{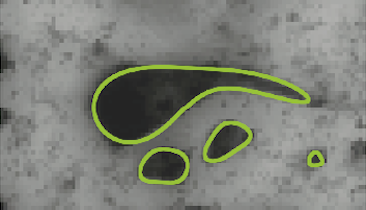}
		\caption{$N=3$, $14$d vector} \label{fig:orderb}
	\end{subfigure}
	\hspace*{\fill}
	\begin{subfigure}{0.32\textwidth}
		\includegraphics[width=\linewidth]{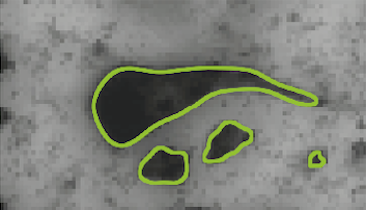}
		\caption{$N=8$, $34$d vector} \label{fig:orderc}
	\end{subfigure}
	\caption{Contour representation with different settings of the order hyperparameter $N$. It defines the vector size of the descriptor that is given by $4N+2$. The higher the order, the more detail is preserved. The 2d contour coordinates are sampled from the descriptor space with \equref{eq:fourier}. Even small settings of $N$ yield good approximations of odd and non-convex shapes, in this case human neuronal cells, including a curved apical dendrite.} \label{fig:order}
\end{figure}

Following \citet{Kuhl1982EllipticFF}, we define a contour of degree $N$ as a series of 2d coordinates $(x_N(t_1), y_N(t_1)), \dots (x_N(t_S), y_N(t_S))$ with $t_s<t_{s+1}$, using the Fourier sine and cosine transformation

\begin{equation} \label{eq:fourier}
    \begin{aligned}
        x_N(t) &= a_0 + \sum^N_{n=1} \br{a_n \sin\br{\frac{2n\pi t}{T}} + b_n \cos\br{\frac{2n\pi t}{T}}}\\
        y_N(t) &= c_0 + \sum^N_{n=1} \br{c_n \sin\br{\frac{2n\pi t}{T}} + d_n \cos\br{\frac{2n\pi t}{T}}}
    \end{aligned}
\end{equation}
For legibility we omit the subscript of $x_N$ and $y_N$ in the following.
The evolution of the x-coordinate along the contour $x(t)$  is parameterized by two series of coefficients $\bm{a}=a_0, a_1,\dots a_N$ and $\bm{b}=b_1,\dots b_N$, with $a_0$ determining the spatial offset of the contour on the pixel grid. Accordingly, $y(t)$ is parameterized by coefficients $\bm{c}$
and $\bm{d}$.
The parameter vector $[\bm{a}, \bm{b}, \bm{c}$ $\bm{d}] \; \in \mathbb{R}^{4N+2}$ hence determines a 2D object contour.
The location parameter $t_s\in [0, 1]$ with interval length $T=1$ determines at which fraction of the contour a coordinate is sampled.
The \emph{order} hyperparameter $N$ determines the smoothness of the contour, with larger $N$ adding higher frequency coefficients and thus allowing closer approximations of object contours (\figref{fig:order}).
This formulation always produces closed contours.
It is differentiable, and both the contour representation and the sampled contour coordinates are fixed in size, given an order $N$ and a sample size $S$.
Thus, we can  directly regress the parameters of this representation to predict closed object contours with convolutional neural networks.

\sloppy The CPN employs two separate regression heads for predicting contour shape $\Set{a_i, b_i, c_i, d_i | 1\leq i\leq N}$ and localization in the image $(a_0,c_0)$. By isolating regression of shape and location, we intend to preserve translational invariance of the contour representation and equivariance of the offset regression.

\subsection{Local Refinement}
\label{sec:refinement}

\begin{figure}
	\centering
	\begin{subfigure}{0.49\linewidth}
		\includegraphics[width=\linewidth]{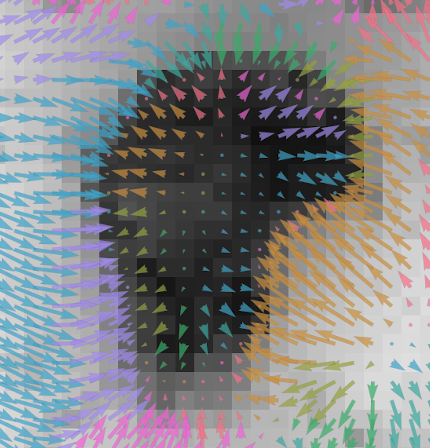}
		\caption{Refinement tensor $\bm{v}$} \label{fig:refinementa}
	\end{subfigure}
	\hspace*{\fill}
	\begin{subfigure}{0.49\linewidth}
		\includegraphics[width=\linewidth]{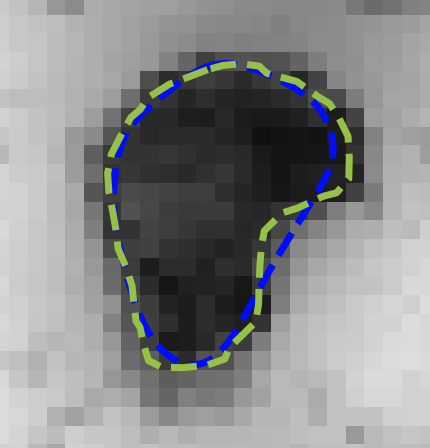}
		\caption{Refinement: before (\emph{blue}), after (\emph{green})} \label{fig:refinementb}
	\end{subfigure}
	\caption{Local refinement example. \ref{fig:refinementa} illustrates the learned refinement tensor $\bm{v}$ as a vector field, superimposed with the input image. \ref{fig:refinementb} shows a contour proposal before and after refinement. The refinement tensor learned to shift contour coordinates to maximize pixel-precision. (Best viewed in color)} \label{fig:refinement}
\end{figure}

To maximize the pixel-precision of estimated contours, we propose a \textit{local refinement} of predicted contour coordinates in the pixel domain as defined in \algoref{alg:refine}.
Using an additional regression head we generate a two channel feature map $\bm{v}$ which represents a 2D residual field on the pixel grid.
We correct each rounded contour proposal coordinate $\lfloor x(t)\rceil, \lfloor y(t)\rceil$ using its residual $\sigma\tanh \left(\bm{v}_{\lfloor x(t)\rceil, \lfloor y(t)\rceil}\right)$, with $\sigma$ as the maximum correction margin, minimizing the distance between estimated and actual contour coordinate.\footnote{Note that the use of rounded coordinates prevents contour proposal heads from being influenced by the refinement head during training.
Also the rounding provides a consistent starting point for the refinement.} 
This correction can be applied multiple times by reusing $\bm{v}$ at updated pixel coordinates.
\figref{fig:refinement} shows an example.
A contour coordinate has reached its final position once $\bm{v}$ yields an offset of zero for all spatial dimensions at a given location.
The combination of correction margin $\sigma$ and the number of iterations limits the influence that the refinement may have on the final result.
The local refinement reduces localization errors of the contour regression and can compensate the exclusion of higher contour frequencies by choosing small values for the order hyperparameter $N$.
Such a refinement becomes tractable since the CPN directly outputs boundary coordinates - in fact the procedure can be efficiently implemented using \textit{fancy indexing}.
The prediction of the refinement tensor $\bm{v}$ is trained implicitly by minimizing the distance between refined contour coordinates and ground truth coordinates in pixel space.

\begin{algorithm}
	\caption[]{Local Refinement. Iteratively refine a contour coordinate $x, y$ using refinement tensor $\bm{v} \in \mathbb{R}^{w \times h \times 2}$ and maximum correction margin $\sigma$, assuming $1\leq x \leq w$ and $1\leq y \leq h$. Rounding is denoted by $\round{\cdot}$.}
	\begin{algorithmic}[1]
		\Procedure{Refine}{x, y, $\bm{v}$, $r$, $\sigma$}
		\For{$r$ iterations}
			\State{
				$\begin{bmatrix} x & y \end{bmatrix} \gets \begin{bmatrix} \round{x} & \round{y} \end{bmatrix} + \sigma\tanh \left(\bm{v}_{\round{x}, \round{y}}\right)$
			}
		\EndFor
		\Return $x,y$
		\EndProcedure
	\end{algorithmic} \label{alg:refine}
\end{algorithm}

\subsection{Non-Maximum Suppression}
Similar to other object detection methods (\cite{he2018mask, redmon2016look, lin2018focal}) the CPN generates dense proposals, thus multiple pixels of the produced output grid may represent the same object.
To remove redundant detections during inference, we apply bounding-box non-maximum suppression (\emph{NMS}).
NMS specifically keeps proposals with a high detection score, but suppresses proposals with lower scores and a bounding box \textit{IoU} (Intersection over Union) that exceeds a given threshold.
As the CPN outputs lists of contour coordinates, we can define bounding boxes very efficiently for each contour proposal $(x(t_1), y(t_1)), \dots (x(t_S), y(t_S))$ as $b = \left[ \min x(\bm{t}), \min y(\bm{t}) , \max x (\bm{t}), \max y(\bm{t})  \right] $.

\subsection{Loss functions}
We define objectives for two components: Detection score and contour prediction.
For legibility we present objectives per pixel.

\paragraph{Detection score}
The detection head performs binary classification for each pixel, producing a score that states whether an object instance is present or not at the pixel location.
The loss $\loss_{\text{inst}}$ for this task is the standard Binary Cross Entropy (\textit{BCE}).

\paragraph{Contour Coordinate Loss}
At each pixel where a contour should be attached, we apply a loss that minimizes the distance between ground truth contour coordinates and estimated coordinates.
For a single coordinate it is given by
\begin{equation}
    \loss_{\text{coord}}(x, y, \hat{x}, \hat{y}) = \frac{1}{2} (\abs{x - \hat{x}} + \abs{y - \Hat{y}})
\end{equation}
The contour proposal prediction is trained using
\begin{equation}
    \loss_{\text{contour}} = \frac{1}{S} \sum_{s=1}^S \loss_{\text{coord}}( x(t_s), y(t_s), \hat{x}(t_s), \Hat{y}(t_s) )
\end{equation}
with ground truth contour coordinate $x(t_s), y(t_s)$ and estimated $\Hat{x}(t_s), \Hat{y}(t_s)$, at random positions $t_s \in [0, 1]$.
Coordinates are defined as in \equref{eq:fourier} given targets $\bm{a}, \bm{b}, \bm{c}, \bm{d}$ and estimates $\bm{\hat{a}}, \bm{\hat{b}}, \bm{\hat{c}}$ and $\bm{\hat{d}}$.
Local refinement is trained accordingly with
\begin{equation}
    \loss_{\text{refine}} = \frac{1}{S} \sum_{s=1}^S \loss_{\text{coord}} \Big( x(t_s), y(t_s), \Call{Refine}{\hat{x}(t_s), \Hat{y}(t_s)} \Big)
\end{equation}
substituting $\Hat{x}(t_s), \Hat{y}(t_s)$ with refined coordinates using \algoref{alg:refine}.

\paragraph{Representation Loss}
Additionally, we can directly supervise the shape parameters in the frequency domain using
\begin{equation}
    \loss_{\text{repr}} = \abs{\bm{\beta}\odot (\bm{a}-\hat{\bm{a}})} + \abs{\bm{\beta}_{-0}\odot (\bm{b}-\hat{\bm{b}})} + \abs{\bm{\beta}\odot (\bm{c}-\hat{\bm{c}})} + \abs{\bm{\beta}_{-0}\odot (\bm{d}-\hat{\bm{d}})}
\end{equation}
with $\bm{\beta}_{-0}$ denoting the exclusion of $\beta_0$ from $\bm{\beta} = [\beta_0, \dots \beta_N]$.
While the objective is already well defined without this representation loss, it provides additional regularization of the shape space and enables to emphasize specific detail levels by applying individual factors $\beta_n$.
An intuitive setting decreases $\beta_n$ with growing $n$ to put more relative emphasis on the coarse contour outlines represented by low frequency coefficients.

\paragraph{CPN Loss}
Combining the components above, the overall per pixel loss is given by
\begin{equation}
    \loss_{\text{CPN}} = \loss_{\text{inst}}(o) + o (\loss_{\text{contour}} + \loss_{\text{refine}} + \lambda \loss_{\text{repr}})
\end{equation}
with $o=1$ for pixels that represent an object and $o=0$ otherwise.

\section{Experiments and results}

We evaluate the instance segmentation performance of the CPN on three datasets (\ncb, \bbbc{39}, \synth{}) and compare the results with U-Net and Mask R-CNN as baseline models.
Also, the cross-dataset generalization performance is examined by training models on \bbbc{39} and testing them on a fourth dataset, \bbbc{41}. To better understand the effects of employing the CPN feature space, this experiment includes a U-Net that is first trained as part of a CPN.
Finally, we compare inference speeds of different models.

\label{experiments}

\begin{figure*}
    \centering
    \includegraphics[width=\textwidth]{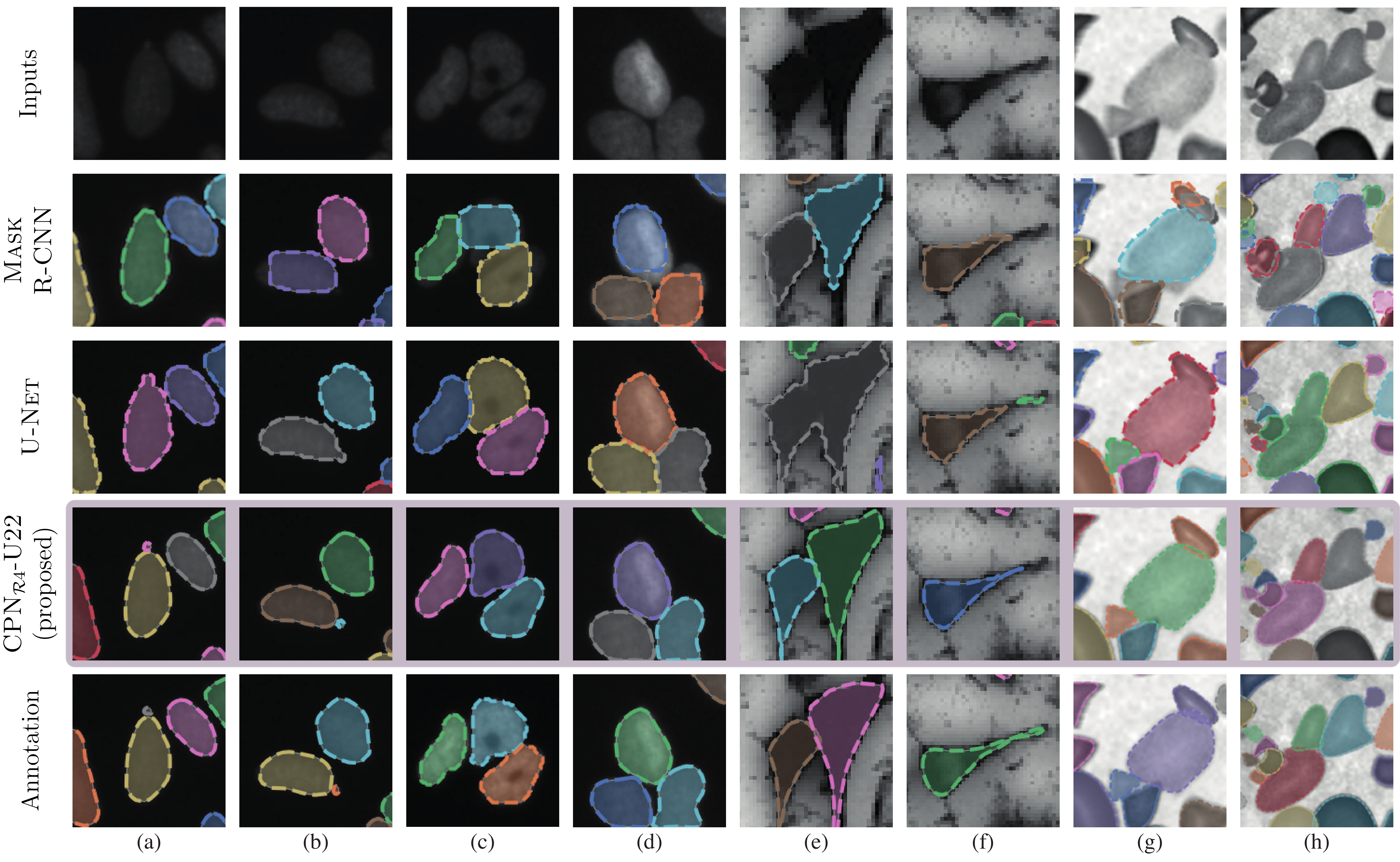}
	\caption{Example patches from different datasets with reference annotations (\nth{5} row) and detections computed by the proposed \cpn{4}{\bbunet} (\nth{4} row), \unet{} (\nth{3} row) and \maskrcnn{} (\nth{2} row) models.}
	\label{fig:systematic_errors}
\end{figure*}

\subsection{Datasets}
\label{sec:datasets}

\begin{figure}
	\centering
	\begin{subfigure}{0.23\textwidth}
		\includegraphics[width=\linewidth]{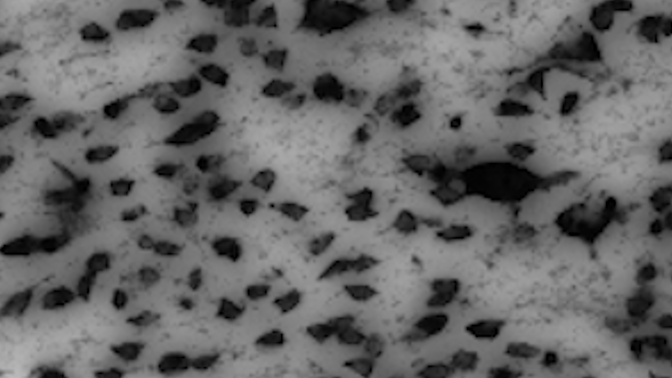}
	\end{subfigure}
	\hspace*{\fill}   
	\begin{subfigure}{0.23\textwidth}
		\includegraphics[width=\linewidth]{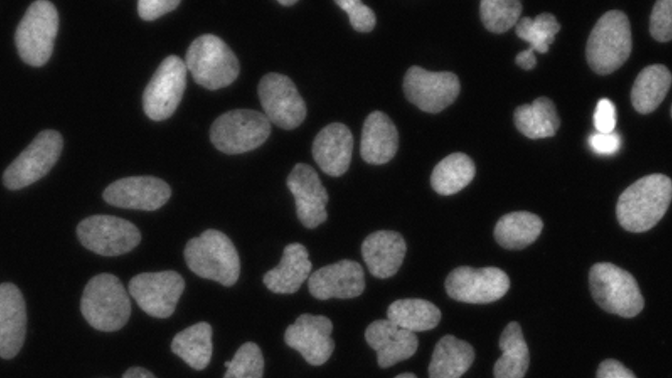}
	\end{subfigure}
	\hspace*{\fill}   
	\begin{subfigure}{0.23\textwidth}
		\includegraphics[width=\linewidth]{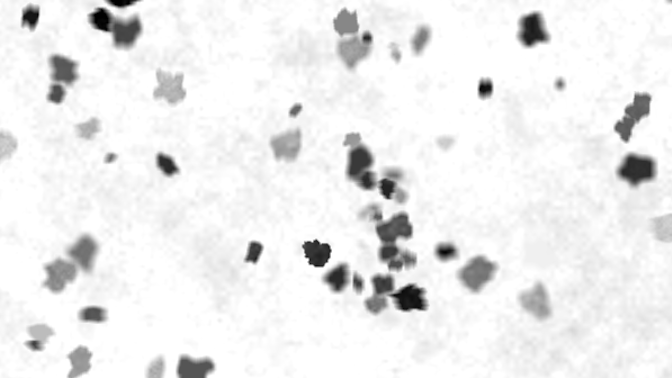}
	\end{subfigure}
	\hspace*{\fill}  
	\begin{subfigure}{0.23\textwidth}
		\includegraphics[width=\linewidth]{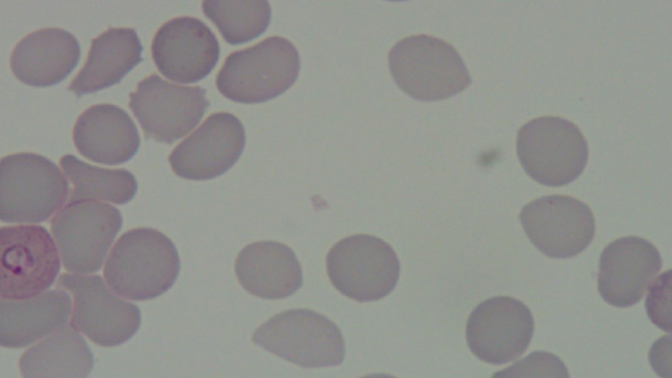}
	\end{subfigure}\\
	
	\begin{subfigure}{0.23\textwidth}
		\includegraphics[width=\linewidth]{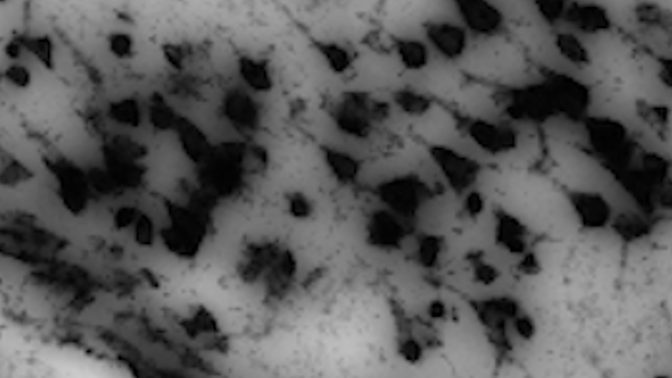}
	\end{subfigure}
	\hspace*{\fill} 
	\begin{subfigure}{0.23\textwidth}
		\includegraphics[width=\linewidth]{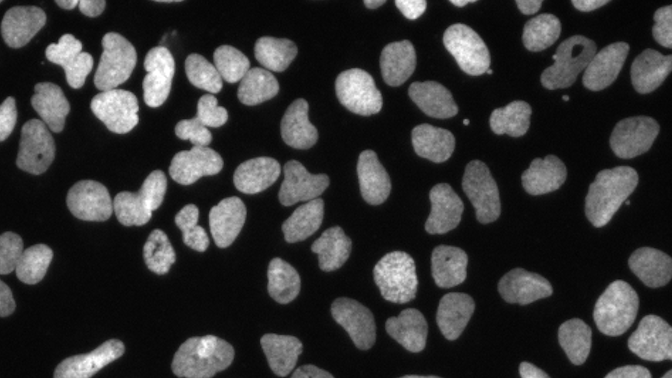}
	\end{subfigure}
	\hspace*{\fill}   
	\begin{subfigure}{0.23\textwidth}
		\includegraphics[width=\linewidth]{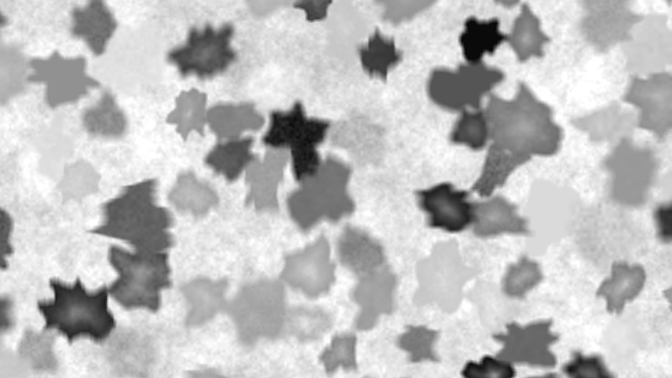}
	\end{subfigure}
	\hspace*{\fill}  
	\begin{subfigure}{0.23\textwidth}
		\includegraphics[width=\linewidth]{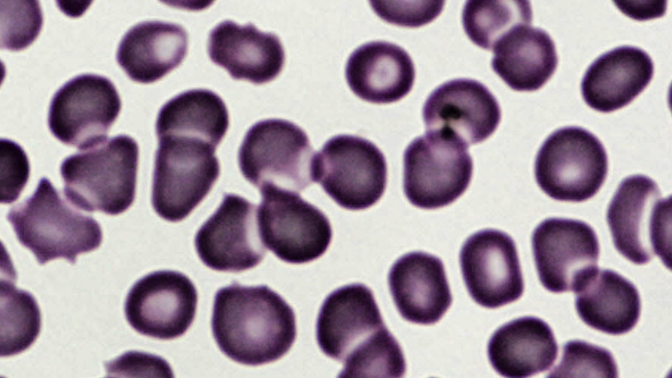}
	\end{subfigure}\\

	\begin{subfigure}{0.23\textwidth}
		\includegraphics[width=\linewidth]{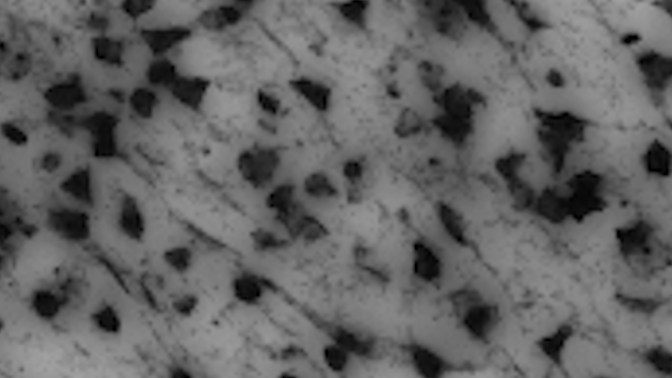}
		\caption{NCB}
		\label{fig:datasets:ncb}
	\end{subfigure}
	\hspace*{\fill} 
	\begin{subfigure}{0.23\textwidth}
		\includegraphics[width=\linewidth]{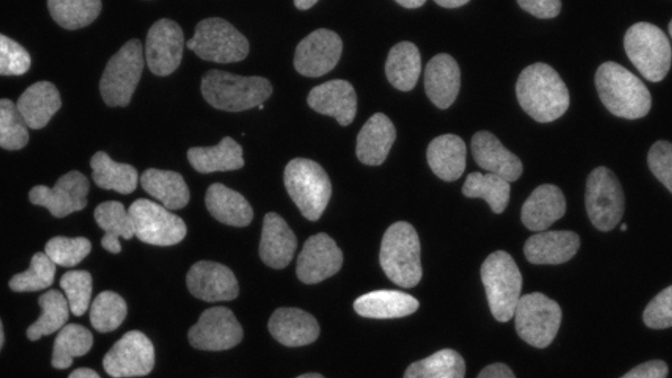}
		\caption{\bbbc{39}}
		\label{fig:datasets:bbbc039}
	\end{subfigure}
	\hspace*{\fill}   
	\begin{subfigure}{0.23\textwidth}
		\includegraphics[width=\linewidth]{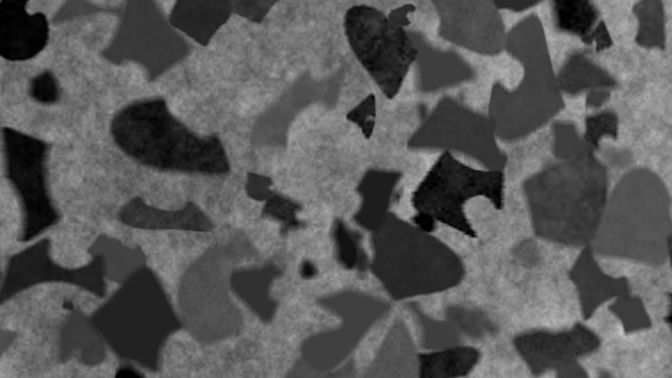}
		\caption{\synth}
		\label{fig:datasets:synth}
	\end{subfigure}
	\hspace*{\fill} 
	\begin{subfigure}{0.23\textwidth}
		\includegraphics[width=\linewidth]{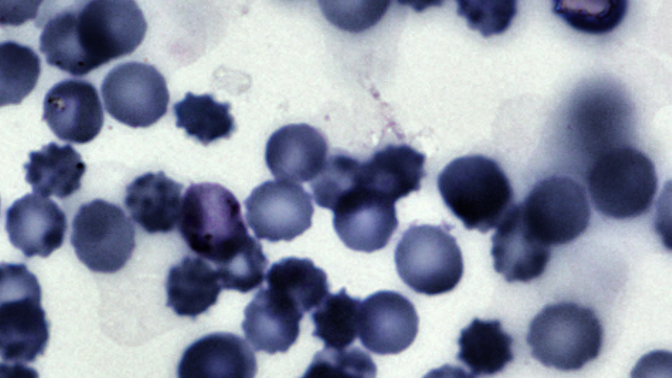}
		\caption{\bbbc{41}}
		\label{fig:datasets:bbbc041}
	\end{subfigure}\\
	\caption{Examples from the used datasets (\secref{sec:datasets}). 
	} \label{fig:datasets}
\end{figure}

\paragraph{NCB - Neuronal Cell Bodies}
This dataset consists of 82 grayscale image patches from microscopic scans of cellbody-stained brain tissue sections, with annotations of approximately 29,000 cell bodies.
\figref{fig:datasets:ncb} shows examples. It includes significant variations in cell shape, intensity, and object overlap,  as well as challenging configurations like occlusions, noise, varying contrast and histological artifacts. Brain samples come from the body donor program of the Anatomical Institute of Düsseldorf in accordance with legal and ethical requirements\footnote{ethics approval \#$4863$}. Tissue sections were stained using a modified Merker stain \citep{merker1983_SilverStaining}. Each tissue section has an approximate thickness of 20 $\text{µm}$ and is captured with a resolution of 1 $\text{µm}$ using a high-throughput light-microscopic scanner
(TissueScope HS, Huron Digital Pathology Inc.).
Note that cell bodies in this dataset are always continuously and fully annotated even under occlusions, in order to allow a model to learn mostly realistic morphologies.
Image patches were manually labeled for cell body instances by a group of experts in our institute. 
This was performed using a custom web-based annotation software, which allowed to enter overlapping pixel labels and to inspect the 3d context provided by depth focusing.
To minimize highly subjective annotations of ambiguous cases, the software includes collaborative feedback features that allow consensus among multiple experts during annotation.
The complete dataset will be publicly available on the EBRAINS\footnote{\url{https://ebrains.eu}} platform.

\paragraph{\bbbc{39} - Nuclei of U2OS cells in a chemical screen}
This is a dataset from the \emph{Broad Institute Bioimage Benchmark Collection}  \citep{ljosa_annotated_2012}. It consists of 200 grayscale images from a high-throughput chemical screen on U2OS cells, depicting  approximately 23,000 annotated nuclei. \figref{fig:datasets:bbbc039} shows examples.

\paragraph{\bbbc{41} - P. vivax (malaria) infected human blood smears}
Also from the Broad Bioimage Benchmark Collection \citep{ljosa_annotated_2012}, this dataset consists of 1364 images depicting approximately $80,000$ Malaria infected human blood smear cells, annotated with bounding boxes.  \figref{fig:datasets:bbbc041} shows examples.

\paragraph{\synth{} - Synthetic shapes} 
This dataset consists of 4129 grayscale images that show a large variety of different synthetic shapes in different sizes. It contains approximately 1,305,000 annotated objects.
\figref{fig:datasets:synth} shows examples.
Shapes include simple structures, such as circles, ellipses and triangles, as well as more complex non-convex structures. Objects and background vary in intensity and texture, with objects showing mostly darker intensities than background.
Similar to the NBC dataset mentioned above, objects can overlap and are fully annotated, even if occluded. Thus a single pixel can belong to more than one instance.

\subsection{Baseline Methods}
\label{sec:baselines}
We evaluate performance against the same baseline methods used in \cite{schmidt2018_CellDetection}.
\begin{itemize}
    \item \unet{} \citep{DBLP:journals/corr/RonnebergerFB15} is an encoder-decoder network with lateral skip-connections and a de-facto standard for biomedical image segmentation (cf.~\secref{sec:related}). In addition to its original definition we use batch normalization after each convolutional layer. Following \cite{Caicedo335216} the network classifies each pixel into one of three classes: cell, background and boundary.
\item \maskrcnn~\citep{he2018mask} is a widely used instance segmentation method that proposes bounding boxes for each object, filters proposals by non maximum suppression and finally produces masks based on proposed bounding box regions (cf.~\secref{sec:related}). The implementation used in our experiments is based on \emph{torchvision}, a Python package that includes popular model architectures and is part of \emph{PyTorch}.
\end{itemize}

\subsection{CPN Training}
\label{sec:cpn-training}
For comparability, we instantiate CPN models with the same backbone architectures as the baseline models, and  train them with the same number of epochs, data size, batch size and data augmentation.
In particular, we use four CPN variants:
\begin{itemize}
    \item 
    \cpn{4}{\bbr{50}{FPN}} uses a Feature Pyramid Network (\emph{FPN}) \citep{lin2017feature} with a $50$ layer residual architecture \citep[ResNet-50]{he2015deep} as its backbone and applies 4 iterations of contour refinement (\secref{sec:refinement})
    \item
    \cpn{0}{\bbr{50}{FPN}} is \cpn{4}{\bbr{50}{FPN}} with contour refinement disabled
    \item 
    \cpn{4}{\bbunet} uses a 22 layer U-Net as a backbone, which is setup like the baseline described in \secref{sec:baselines}, but omitting its final output layer. It uses 4 iterations of contour refinement
    \item
    \cpn{0}{\bbunet} is  \cpn{4}{\bbunet} with contour refinement disabled
\end{itemize}
For assessing inference speed, we will use additional backbone architectures (\secref{sec:inference speed}). 

We supervise both the contour representation and the sampled contour coordinates.
As the contour representation is well defined, we calculate ground truth representations on the fly and use them to guide the network during training.
Using \equref{eq:fourier} with uniform sampling $t_1, \dots t_S$ ($t_s\in [0,1]$) we retrieve contour coordinates from both ground truth and prediction for supervision. The sample size hyperparameter $S$ influences precision and performance by fixing the number of coordinates used during training. We choose $S=64$ here.

While it is also possible to use the derivable and non-parametric formula from \citet{Kuhl1982EllipticFF} to derive the contour representation from another latent space, we did not observe any benefits and thus omit this possibility.

\begin{table*}[t]
	\caption{
	Instance segmentation results for selected datasets and methods. The \textit{F1 score} F1$_{\tau=0.60}$ is reported for a range of \textit{intersection over union} (IoU) thresholds $\tau$ and as the average F1$_{\text{avg}} = 1/9\sum_{\tau\in \mathrm{T}} \text{F1}_{\tau}$ for thresholds $\mathrm{T} = (0.5, 0.55, 0.6, \dots 0.9)$.}
	\label{tab:results}
	\centering
	\begin{tabular}{llc|ccccc}
		\toprule
		Model & Backbone & F1$_{\text{avg}}$ & \fscore{0.5} & \fscore{0.6} & \fscore{0.7} & \fscore{0.8} & \fscore{0.9}  \\
		\toprule
		\multicolumn{8}{c}{Neuronal Cell Bodies}                   \\
		\midrule
			\cpn{4}{\bbunet} & U-Net & $\mathbf{0.55}$ & $\mathbf{0.80}$ & $\mathbf{0.74}$ & $\mathbf{0.62}$ & $\mathbf{0.40}$ & $\mathbf{0.10}$\\
			\cpn{0}{\bbunet} & U-Net & ${0.51}$ & $\mathbf{0.80}$ & ${0.73}$ & ${0.58}$ & ${0.33}$ & ${0.05}$\\
			\cpn{4}{\bbr{50}{FPN}} & ResNet-50-FPN & ${0.43}$ & ${0.74}$ & ${0.65}$ & ${0.49}$ & ${0.23}$ & ${0.02}$\\
			\cpn{0}{\bbr{50}{FPN}} & ResNet-50-FPN & ${0.42}$ & ${0.73}$ & ${0.64}$ & ${0.48}$ & ${0.22}$ & ${0.02}$\\
			\unet & U-Net & ${0.47}$ & ${0.71}$ & ${0.63}$ & ${0.51}$ & ${0.33}$ & $\mathbf{0.10}$\\
			Mask R-CNN & ResNet-50-FPN & ${0.34}$ & ${0.70}$ & ${0.55}$ & ${0.34}$ & ${0.11}$ & ${0.00}$\\
		\toprule
		\multicolumn{8}{c}{\bbbc{39}}                   \\
		\midrule
			\cpn{4}{\bbunet} & U-Net & $\mathbf{0.91}$ & $\mathbf{0.96}$ & $\mathbf{0.95}$ & $\mathbf{0.93}$ & $\mathbf{0.91}$ & $\mathbf{0.76}$\\
			\cpn{0}{\bbunet}  & U-Net & ${0.90}$ & $\mathbf{0.96}$ & $\mathbf{0.95}$ & $\mathbf{0.93}$ & ${0.90}$ & ${0.72}$\\
			\cpn{4}{\bbr{50}{FPN}} & ResNet-50-FPN & ${0.90}$ & ${0.95}$ & ${0.94}$ & $\mathbf{0.93}$ & ${0.90}$ & ${0.74}$\\
			\cpn{0}{\bbr{50}{FPN}} & ResNet-50-FPN & ${0.90}$ & ${0.95}$ & ${0.94}$ & $\mathbf{0.93}$ & ${0.89}$ & ${0.71}$\\
			\unet & U-Net & ${0.89}$ & ${0.95}$ & ${0.93}$ & ${0.92}$ & ${0.88}$ & ${0.71}$\\
			\maskrcnn & ResNet-50-FPN & ${0.86}$ & ${0.94}$ & ${0.93}$ & ${0.92}$ & ${0.89}$ & ${0.52}$\\
		\toprule
		\multicolumn{8}{c}{Synthetic Shapes}                   \\
		\midrule
			\cpn{4}{\bbunet} & U-Net & $\mathbf{0.90}$ & $\mathbf{0.98}$ & $\mathbf{0.98}$ & $\mathbf{0.96}$ & $\mathbf{0.89}$ & $\mathbf{0.64}$\\
			\cpn{0}{\bbunet} & U-Net  & ${0.89}$ & $\mathbf{0.98}$ & $\mathbf{0.98}$ & $\mathbf{0.96}$ & ${0.88}$ & ${0.51}$\\
			\cpn{4}{\bbr{50}{FPN}}  & ResNet-50-FPN & ${0.88}$ & $\mathbf{0.98}$ & ${0.97}$ & ${0.94}$ & ${0.86}$ & ${0.54}$\\
			\cpn{0}{\bbr{50}{FPN}}  & ResNet-50-FPN & ${0.86}$ & $\mathbf{0.98}$ & ${0.97}$ & ${0.94}$ & ${0.84}$ & ${0.47}$\\
			\unet & U-Net & ${0.87}$ & ${0.96}$ & ${0.95}$ & ${0.92}$ & ${0.85}$ & ${0.59}$\\
			\maskrcnn & ResNet-50-FPN & $0.85$ & $0.96$ & $0.90$ & $0.85$ & $0.72$ & $0.36$ \\
		\bottomrule
	\end{tabular}
\end{table*}

\begin{table*}
	\caption{Cross-dataset evaluation of object detection performance. We report F1 scores for models trained on \bbbc{39} dataset and tested on \bbbc{41} dataset. Results are based on bounding boxes using same metrics as \tabref{tab:results}. The pretrained U-Net was trained as part of \cpn{4}{\bbunet}.}
	\label{tab:domain_shift_results}
	\centering
	\begin{tabular}{llc|ccccc}
		\toprule
		Model & Backbone & F1$_{\text{avg}}$ & \fscore{0.5} & \fscore{0.6} & \fscore{0.7} & \fscore{0.8} & \fscore{0.9} \\
		\midrule
			\cpn{4}{\bbunet} & U-Net & $\bm{0.54}$  & $\bm{0.83}$  & $\bm{0.81$}  & $\bm{0.70}$  & ${0.29}$  & $0.02$ \\
			\unet & Pretrained U-Net & $0.52$  & $0.72$  & $0.71$  & $0.67$  & $\bm{0.39}$  & $\bm{0.07}$ \\
			\unet & U-Net & $0.45$  & $0.62$  & $0.60$  & $0.57$  & $0.33$  & $0.06$ \\
			\maskrcnn & ResNet-50-FPN & $0.49$  & $0.77$  & $0.75$  & $0.66$  & $0.24$  & $0.02$ \\
		\bottomrule
	\end{tabular}
\end{table*}

\subsection{Detection and segmentation performance}
To evaluate the detection performance and the shape quality of the produced contours we use the harmonic mean of precision and recall $\text{F1}_\tau=\frac{TP_\tau}{TP_\tau+\frac{1}{2}(FP_\tau+FN_\tau)}$ for different \emph{Intersection over Union} (IoU) thresholds $\tau$.
The IoU threshold $\tau \in[0,1]$ defines the minimal IoU that is required for two shapes to be counted as a match. 
Each ground truth shape can be a match for at most one predicted shape.
A True Positive $(\text{TP})$ is a predicted shape that matches a ground truth shape, a False Positive ($\text{FP}$) is a shape that does not match any ground truth shape and a False Negative ($\text{FN}$) is a ground truth shape that does not match any predicted shape.
$\text{F1}_\tau$ scores with a small $\tau=0.5$ quantify the coarse detection performance of a model, yielding good scores if the model correctly infers object presence along with a roughly matching contour.
$\text{F1}_\tau$ scores with a larger $\tau=0.9$ quantify the fine detection performance, allowing little deviance from the target shape.
We define F1$_{\text{avg}} = 1/9\sum_{\tau\in \mathrm{T}} \text{F1}_{\tau}$ for thresholds $\mathrm{T} = (0.5, 0.55, 0.6, \dots 0.9)$, to measure the average performance for different thresholds.

\tabref{tab:results} shows quantitative results of the CPN, U-Net and Mask R-CNN on three different datasets.
The CPN with local refinement yields highest scores on all datasets.
Local refinement further increases the average F1 scores, especially for high thresholds $\tau$, thus increasing the quality of the contours as expected. On the datasets \bbbc{39} and \synth{} \cpn{4}{\bbunet} outperforms the baseline models for all thresholds.

\subsection{Cross-dataset generalization}

We assessed how well the baseline and CPN models generalize to variations in the input data distribution as follows:
Models are trained for instance segmentation on the \bbbc{39} dataset. Without any retraining or adaptation, models are then applied to \bbbc{41}. Generalization capabilities are then evaluated with the F1 score on the basis of bounding boxes derived from the respective instance segmentation results.
This provides a quantitative characterization of detection and segmentation performance under transfer to different data domains.
To comply with the basic characteristics of \bbbc{39}, we converted the images to inverted grayscale images and applied a fixed contrast adjustment and downscaling.

\begin{figure*}
	\centering
	  \begin{tikzpicture}[node distance=10.6em]
        \node (gt1) at (0,0) {\includegraphics[width=10.3em]{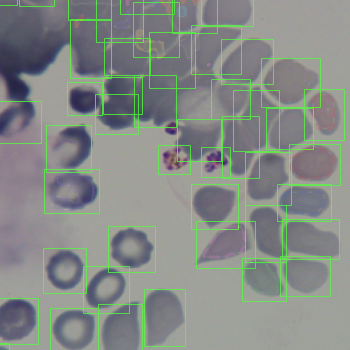}};
        \node[right of=gt1] (cpn1) {\includegraphics[width=10.3em]{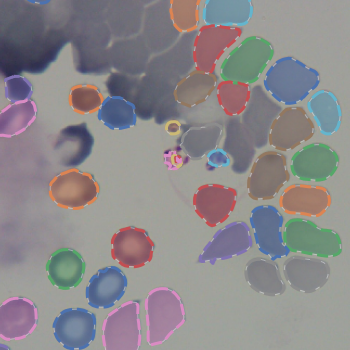}};
        \node[right of=cpn1] (unet1) {\includegraphics[width=10.3em]{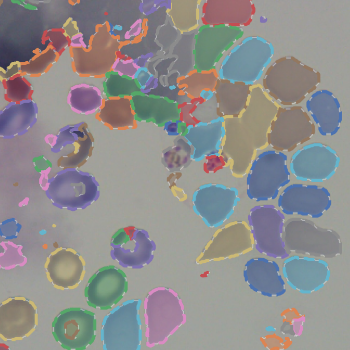}};
        \node[right of=unet1] (mrcnn1) {\includegraphics[width=10.3em]{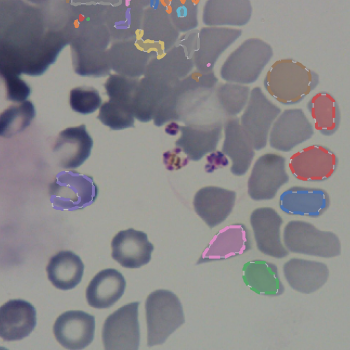}};   
    
        \node[below of=gt1] (gt2) {\includegraphics[width=10.3em]{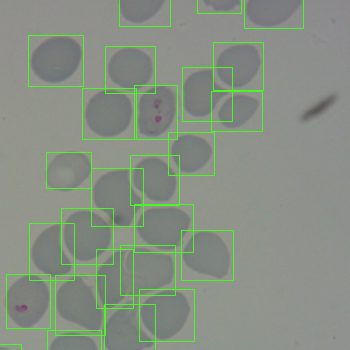}}; 
        \node[right of=gt2] (cpn2) {\includegraphics[width=10.3em]{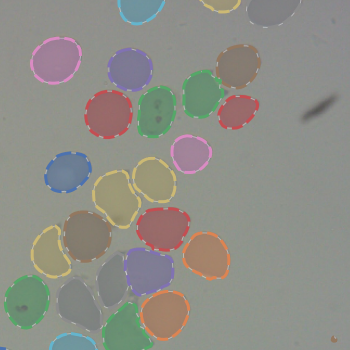}};
        \node[right of=cpn2] (unet2) {\includegraphics[width=10.3em]{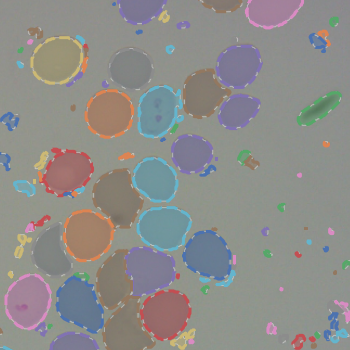}};
        \node[right of=unet2] (mrcnn2) {\includegraphics[width=10.3em]{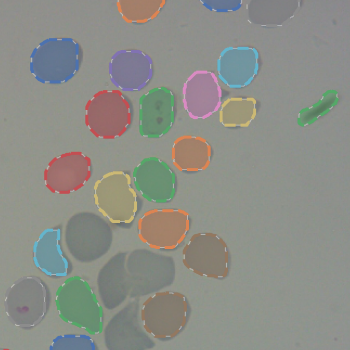}};

        \node[below of=gt2,node distance=6.15em] (gtl) {\footnotesize (a) Annotation};
        \node[right of=gtl] (cpnl) {\footnotesize (b) \cpn{4}{\bbunet}};
        \node[right of=cpnl] (unetl) {\footnotesize (c) \unet};
        \node[right of=unetl] (maskrcnnl) {\footnotesize (d) \maskrcnn};
    \end{tikzpicture}
    
	\caption{Cross-dataset generalization examples from three different models. The models were trained on \bbbc{39} and applied to images from \bbbc{41} without retraining or adaptation. Two samples are depicted above.
	} \label{fig:cross-data}
\end{figure*}
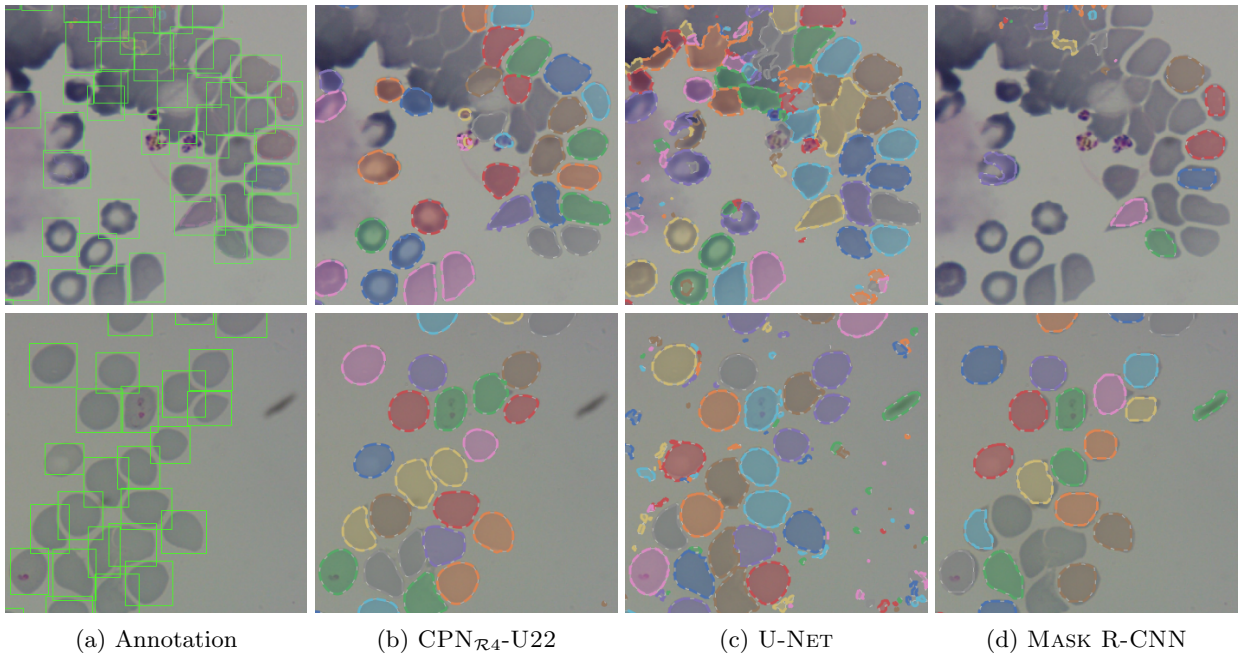

Results are shown in \tabref{tab:domain_shift_results}.
CPN models consistently show higher scores than baseline methods. For small IoU thresholds such as $\tau=0.50$, the scores of CPN and Mask R-CNN models are particularly distinguished from U-Net.
\figref{fig:cross-data} shows detected instances from different methods on two typical examples, illustrating the problems in cross-dataset generalization.
\cpn{4}{\bbunet} tends to detect conservatively, preferring some false negatives for avoiding false positives. \unet{} shows more false positives, sometimes seemingly detecting noise. For \maskrcnn, contours are less precise than for the others, and overall less true instances are detected.

\paragraph{Reusing trained CPN backbones for different tasks}
We also examined the generalization performance of a \unet{} when its encoder and decoder are trained as the backbone of \cpn{4}{\bbunet}, and then reused with a new final prediction layer in a \unet{} to output segmentation results. For retraining, the encoder part is frozen, to ensure that the CPN feature space is kept intact.
This case is reported in the second row of \tabref{tab:domain_shift_results}. All scores are significantly higher compared to the \unet{} without such pre-training.  
For high IoU thresholds, e.g. $\tau \in \{0.8, 0.9\}$, this variant even provides overall highest scores.
However, as the generalization performance increased, the F1$_{\text{avg}}$ score on the \bbbc{39} test set dropped from $0.89$ to $0.87$. 

\subsection{Inference speed}
\label{sec:inference speed}

\begin{table}[t]
	\caption{Inference speeds of different models. We report the number of frames per second (\textit{FPS}) for the \bbbc{39} test set with an image size of $520\times 696$. We measure times with single-precision (\emph{float32}) and and automatic mixed precision (\emph{amp}). The initial run and possible post-processing steps are excluded. All models were implemented and executed as PyTorch models on an NVIDIA A100. We denote ResNet by 'R', ResNeXt by 'X' and U-Net by 'U' for brevity.
	}
	\label{tab:times}
	\centering
	\begin{tabular}{lcc}
		\toprule
		Model & FPS & FPS (amp) \\
		\midrule
			\cpn{0}{\bbr{50}{FPN}} & $30.19$ & $37.57$ \\
			\cpn{4}{\bbr{50}{FPN}} & $29.86$ & $36.17$ \\
			\cpn{4}{\bbx{50}{FPN}} & $27.20$ & $36.83$ \\
			\unet & $23.42$ & $77.71$ \\
			CPN$_{\mathcal{R}4}$-U22 ($P_2$ stride 2) & $15.41$ & $42.20$ \\
			\mrcnn{\bbr{50}{FPN}} & $13.74$ & - \\
			\cpn{4}{\bbx{101}{FPN}} & $13.39$ & $25.66$ \\
			CPN$_{\mathcal{R}4}$-U22 & $12.71$ & $26.72$ \\
		\bottomrule
	\end{tabular}
\end{table}

We computed the number of frames per seconds (\emph{FPS}) on the \bbbc{39} test set for different models. Each image has a size of $520\times 696$. To improve the precision of the measurement we reiterated over the test set multiple times.
Pre- and post-processing steps were excluded from the timings, as well as initial warm-up runs.
This experiment was performed in single-precision (\emph{float32}) and automatic mixed precision (\emph{amp}) via PyTorch's \emph{autocast} feature.
The latter automatically selects CUDA operations to run in half-precision (\emph{float16}) to improve performance while aiming to maintain accuracy.

Results are presented in \tabref{tab:times}.
In terms of inference speed, \cpn{4}{\bbr{50}{FPN}} outperforms both \mrcnn{\bbr{50}{FPN}} and \unet{} when applied with normal single-precision (\emph{float32}). As this CPN reached $29.9$ FPS, it qualifies for many online video processing applications.
When applied with automatic mixed precision (\emph{amp}) the \cpn{4}{\bbunet}, that uses a stride of $2$ in the classification and regression head, achieved $42.2$ FPS - the highest performance of the tested CPN models and the second highest overall performance among the tested models.
\unet{}, which shares the same backbone, showed the best inference speed performance using amp.

The influence of local refinement on inference speed was evaluated for the \bbr{50}{FPN} based CPN, for which four refinement iterations reduced the result by 0.33 FPS, when used with single-precision (\emph{float32}).

\section{Discussion and conclusion}
\label{conclusion}

We proposed the Contour Proposal Network (CPN), a framework for segmenting object instances by proposing contours which are encoded as interpretable, fixed-sized representations based on Fourier Descriptors. CPN models can be constructed with different backbone CNN architectures to produce image features.
We assessed the performance of four different CPN variants, employing both U-Net and ResNet-FPN backbones, against a standard U-Net and Mask R-CNN as baseline.
All U-Net based CPNs outperformed the \unet{} counterpart in terms of F1$_{\text{avg}}$ instance segmentation performance on all three tested datasets, both with and without local refinement. Given that CPN and \unet{} share the same backbone architecture \bbunet, the results indicate that the CPN provides a more effective problem description.
This is also supported by the comparison of the CPN and Mask R-CNN in our experiments. For both tested backbone architectures, the CPNs show consistently higher F1$_{\text{avg}}$ than \mrcnn{\bbr{50}{FPN}}.

The CPN models employ a highly entangled representation of object shapes and sparse detections: The networks are effectively forced to concentrate the
description of a complete object into a single pixel by anchoring the boundary representation to a specific coordinate. This requires them to form an intrinsic spatial relationship between whole objects and their parts, which encourages compact and robust representations with good generalization properties. This principle shares some commonalities with \emph{Capsule Networks} \citep{sabourDynamicRoutingCapsules2017}, which also aim to condense instances of objects or object parts into vector representations coupled with a detection score.
The effects can be observed when looking at examples as depicted in \figref{fig:systematic_errors}: If boundaries are invisible or poorly defined, CPN models exploit the learned knowledge of boundary shapes to find highly plausible separations (e.g.~\figref{fig:systematic_errors} e, g, h). Very small and touching objects, which are often overseen by pixel-based methods, are well detected (e.g.~\figref{fig:systematic_errors} a, b). Separation of clusters of kissing objects is typically modelled quite accurately, reproducing gaps between touching shapes very consistently (e.g.~\figref{fig:systematic_errors} c, d). Thin structures, such as dendrites, can be modeled accurately (e.g.~\figref{fig:systematic_errors}f). Discontinuities that may occur with pixel-based methods can be avoided, as proposed contours are continuous and closed by design.

While leading to accurate object representations, experiments on cross-dataset generalization showed that the learned shape priors are not overly restrictive and transfer well to different data distributions. 
Even more, the CPN models are able to produce plausible contours for previously unseen objects as long as their basic morphology is consistent with the training examples.
In particular the F$1_{\tau=0.50}$ margin between \cpn{4}{\bbunet} and \unet{} of $0.21$ suggests that the better performing CPN formed a more universal intrinsic understanding of what an instance is.
In this context, we also observed that the CPN and its objective have a positive influence on the backbone CNNs to produce a feature space with good generalization properties - a pixel-classifying U-Net showed significantly better performance on our cross-dataset evaluation when its encoder and decoder were trained as a CPN backbone.

By modeling the contour representation in the frequency domain, CPNs can bypass several sampling problems occurring in previous works, like selecting the optimal sampling rate in pixel space.
Instead, by setting the order of the Fourier series, the user can specify different levels of contour complexity in a natural way.
Furthermore, the representation allows to generate arbitrary output resolutions without compromising detection accuracy.

The local refinement step, which is an integrated and fully trainable part of the CPN, supports contour proposals to achieve high pixel precision despite the regularization imposed by the shape model. We measured a notable increase in performance for high IoU thresholds when applying refinement, indicating that high contour frequencies can be modeled efficiently using a residual field. While the refinement can improve contour details, it only had a minor influence on inference speed in our experiment. We see the refinement as an important complementary module of the CPN framework.

In terms of inference speed, \cpn{4}{\bbr{50}{FPN}} outperforms all other tested methods when applied with normal single-precision (\emph{float32}). With $29.9$ FPS it is even suitable for online processing tasks, especially as it produces ready-to-use object instance descriptions, not requiring additional post-processing steps like connected component labeling.
The experiments also showed that local refinement adds little time overhead, in the case of \bbr{50}{FPN} based CPN four refinement iterations increased pixel-precision while costing less than half a frame per second.
For automatic mixed precision \cpn{4}{\bbunet} with strided heads showed fasted inference speed among all CPNs with $42.2$ FPS.

Since the only assumption of the proposed approach are closed object contours, it is applicable to a wide range of detection problems, also outside the biomedical domain that have not been investigated in the present work.

An implementation of the model architecture in PyTorch is available at \url{https://github.com/FZJ-INM1-BDA/celldetection}.

%% file: acknowledgments.tex
\section*{Acknowledgments}
This project received funding from the European Union’s Horizon 2020 Research and Innovation Programme, grant agreement 945539 (HBP SGA3), and Priority Program 2041 (SPP 2041) "Computational Connectomics" of the German Research Foundation (DFG). Computing time was granted through JARA-HPC on the supercomputer JURECA at Juelich Supercomputing Centre (JSC) as part of the project CJINM14.

%% file: main.bbl
\begin{thebibliography}{30}
\providecommand{\natexlab}[1]{#1}
\providecommand{\url}[1]{\texttt{#1}}
\expandafter\ifx\csname urlstyle\endcsname\relax
  \providecommand{\doi}[1]{doi: #1}\else
  \providecommand{\doi}{doi: \begingroup \urlstyle{rm}\Url}\fi

\bibitem[Bochkovskiy et~al.(2020)Bochkovskiy, Wang, and
  Liao]{bochkovskiy2020yolov4}
A.~Bochkovskiy, C.-Y. Wang, and H.-Y.~M. Liao.
\newblock Yolov4: Optimal speed and accuracy of object detection, 2020.

\bibitem[Caicedo et~al.(2019)Caicedo, Roth, Goodman, Becker, Karhohs, Broisin,
  Molnar, McQuin, Singh, Theis, and et~al.]{Caicedo335216}
J.~C. Caicedo, J.~Roth, A.~Goodman, T.~Becker, K.~W. Karhohs, M.~Broisin,
  C.~Molnar, C.~McQuin, S.~Singh, F.~J. Theis, and et~al.
\newblock Evaluation of deep learning strategies for nucleus segmentation in
  fluorescence images.
\newblock \emph{Cytometry Part A}, 95\penalty0 (9):\penalty0 952–965, Sep
  2019.
\newblock ISSN 1552-4922, 1552-4930.
\newblock \doi{10.1002/cyto.a.23863}.

\bibitem[Chen et~al.(2016)Chen, Qi, Yu, and Heng]{chen2016dcan}
H.~Chen, X.~Qi, L.~Yu, and P.-A. Heng.
\newblock Dcan: Deep contour-aware networks for accurate gland segmentation,
  2016.

\bibitem[Dietler et~al.(2020)Dietler, Minder, Gligorovski, Economou, Joly,
  Sadeghi, Chan, Koziński, Weigert, Bitbol, and
  Rahi]{dietler_convolutional_2020}
N.~Dietler, M.~Minder, V.~Gligorovski, A.~M. Economou, D.~A. H.~L. Joly,
  A.~Sadeghi, C.~H.~M. Chan, M.~Koziński, M.~Weigert, A.-F. Bitbol, and S.~J.
  Rahi.
\newblock A convolutional neural network segments yeast microscopy images with
  high accuracy.
\newblock \emph{Nature Communications}, 11\penalty0 (1):\penalty0 5723, Dec.
  2020.
\newblock ISSN 2041-1723.
\newblock \doi{10.1038/s41467-020-19557-4}.
\newblock URL \url{http://www.nature.com/articles/s41467-020-19557-4}.

\bibitem[Guerrero-Pena et~al.(2018)Guerrero-Pena, Marrero~Fernandez, Ing~Ren,
  Yui, Rothenberg, and Cunha]{Guerrero_Pena_2018}
F.~A. Guerrero-Pena, P.~D. Marrero~Fernandez, T.~Ing~Ren, M.~Yui,
  E.~Rothenberg, and A.~Cunha.
\newblock Multiclass weighted loss for instance segmentation of cluttered
  cells.
\newblock \emph{2018 25th IEEE International Conference on Image Processing
  (ICIP)}, Oct 2018.
\newblock \doi{10.1109/icip.2018.8451187}.
\newblock URL \url{http://dx.doi.org/10.1109/ICIP.2018.8451187}.

\bibitem[Gur et~al.(2019)Gur, Shaharabany, and Wolf]{gur2019_EndEnd}
S.~Gur, T.~Shaharabany, and L.~Wolf.
\newblock End to end trainable active contours via differentiable rendering.
\newblock \emph{arXiv:1912.00367 [cs]}, Dec 2019.
\newblock URL \url{http://arxiv.org/abs/1912.00367}.
\newblock arXiv: 1912.00367.

\bibitem[{He} et~al.(2016){He}, {Zhang}, {Ren}, and {Sun}]{he2015deep}
K.~{He}, X.~{Zhang}, S.~{Ren}, and J.~{Sun}.
\newblock Deep residual learning for image recognition.
\newblock In \emph{2016 IEEE Conference on Computer Vision and Pattern
  Recognition (CVPR)}, pages 770--778, June 2016.
\newblock \doi{10.1109/CVPR.2016.90}.

\bibitem[He et~al.(2017)He, Gkioxari, Dollar, and Girshick]{he2018mask}
K.~He, G.~Gkioxari, P.~Dollar, and R.~Girshick.
\newblock Mask r-cnn.
\newblock In \emph{2017 IEEE International Conference on Computer Vision
  (ICCV)}, page 2980–2988. IEEE, Oct 2017.
\newblock ISBN 978-1-5386-1032-9.
\newblock \doi{10.1109/ICCV.2017.322}.
\newblock URL \url{http://ieeexplore.ieee.org/document/8237584/}.

\bibitem[Jetley et~al.(2017)Jetley, Sapienza, Golodetz, and
  Torr]{jetley2017straight}
S.~Jetley, M.~Sapienza, S.~Golodetz, and P.~H.~S. Torr.
\newblock Straight to shapes: Real-time detection of encoded shapes.
\newblock In \emph{2017 IEEE Conference on Computer Vision and Pattern
  Recognition (CVPR)}, page 4207–4216. IEEE, Jul 2017.
\newblock ISBN 978-1-5386-0457-1.
\newblock \doi{10.1109/CVPR.2017.448}.
\newblock URL \url{http://ieeexplore.ieee.org/document/8099931/}.

\bibitem[Kuhl and Giardina(1982)]{Kuhl1982EllipticFF}
F.~P. Kuhl and C.~R. Giardina.
\newblock Elliptic fourier features of a closed contour.
\newblock \emph{Computer Graphics and Image Processing}, 18:\penalty0 236--258,
  1982.

\bibitem[Lin et~al.(2014)Lin, Maire, Belongie, Hays, Perona, Ramanan,
  Doll{\'a}r, and Zitnick]{lin2015microsoft}
T.-Y. Lin, M.~Maire, S.~Belongie, J.~Hays, P.~Perona, D.~Ramanan,
  P.~Doll{\'a}r, and C.~L. Zitnick.
\newblock Microsoft coco: Common objects in context.
\newblock In D.~Fleet, T.~Pajdla, B.~Schiele, and T.~Tuytelaars, editors,
  \emph{Computer Vision -- ECCV 2014}, pages 740--755, Cham, 2014. Springer
  International Publishing.
\newblock ISBN 978-3-319-10602-1.

\bibitem[Lin et~al.(2017{\natexlab{a}})Lin, Dollar, Girshick, He, Hariharan,
  and Belongie]{lin2017feature}
T.-Y. Lin, P.~Dollar, R.~Girshick, K.~He, B.~Hariharan, and S.~Belongie.
\newblock Feature {Pyramid} {Networks} for {Object} {Detection}.
\newblock In \emph{2017 {IEEE} {Conference} on {Computer} {Vision} and
  {Pattern} {Recognition} ({CVPR})}, pages 936--944, Honolulu, HI, July
  2017{\natexlab{a}}. IEEE.
\newblock ISBN 978-1-5386-0457-1.
\newblock \doi{10.1109/CVPR.2017.106}.
\newblock URL \url{http://ieeexplore.ieee.org/document/8099589/}.

\bibitem[Lin et~al.(2017{\natexlab{b}})Lin, Goyal, Girshick, He, and
  Dollar]{lin2018focal}
T.-Y. Lin, P.~Goyal, R.~Girshick, K.~He, and P.~Dollar.
\newblock Focal loss for dense object detection.
\newblock In \emph{2017 IEEE International Conference on Computer Vision
  (ICCV)}, page 2999–3007. IEEE, Oct 2017{\natexlab{b}}.
\newblock ISBN 978-1-5386-1032-9.
\newblock \doi{10.1109/ICCV.2017.324}.
\newblock URL \url{http://ieeexplore.ieee.org/document/8237586/}.

\bibitem[Liu et~al.(2016)Liu, Anguelov, Erhan, Szegedy, Reed, Fu, and
  Berg]{Liu_Anguelov_Erhan_Szegedy_Reed_Fu_Berg_2016}
W.~Liu, D.~Anguelov, D.~Erhan, C.~Szegedy, S.~Reed, C.-Y. Fu, and A.~C. Berg.
\newblock \emph{SSD: Single Shot MultiBox Detector}, volume 9905 of
  \emph{Lecture Notes in Computer Science}, page 21–37.
\newblock Springer International Publishing, 2016.
\newblock ISBN 978-3-319-46447-3.
\newblock \doi{10.1007/978-3-319-46448-0_2}.
\newblock URL \url{http://link.springer.com/10.1007/978-3-319-46448-0_2}.

\bibitem[Ljosa et~al.(2012)Ljosa, Sokolnicki, and
  Carpenter]{ljosa_annotated_2012}
V.~Ljosa, K.~L. Sokolnicki, and A.~E. Carpenter.
\newblock Annotated high-throughput microscopy image sets for validation.
\newblock \emph{Nature Methods}, 9\penalty0 (7):\penalty0 637--637, July 2012.
\newblock ISSN 1548-7091, 1548-7105.
\newblock \doi{10.1038/nmeth.2083}.
\newblock URL \url{http://www.nature.com/articles/nmeth.2083}.

\bibitem[Merker(1983)]{merker1983_SilverStaining}
B.~Merker.
\newblock Silver staining of cell bodies by means of physical development.
\newblock \emph{Journal of Neuroscience Methods}, 9\penalty0 (3):\penalty0
  235--241, Nov. 1983.
\newblock ISSN 0165-0270.
\newblock \doi{10.1016/0165-0270(83)90086-9}.

\bibitem[Miksys et~al.(2019)Miksys, Jetley, Sapienza, Golodetz, and
  Torr]{miksys2019straight}
L.~Miksys, S.~Jetley, M.~Sapienza, S.~Golodetz, and P.~H.~S. Torr.
\newblock Straight to shapes++: Real-time instance segmentation made more
  accurate, 2019.

\bibitem[Redmon and Farhadi(2018)]{redmon2018yolov3}
J.~Redmon and A.~Farhadi.
\newblock Yolov3: An incremental improvement, 2018.

\bibitem[Redmon et~al.(2016)Redmon, Divvala, Girshick, and
  Farhadi]{redmon2016look}
J.~Redmon, S.~Divvala, R.~Girshick, and A.~Farhadi.
\newblock You only look once: Unified, real-time object detection.
\newblock In \emph{2016 IEEE Conference on Computer Vision and Pattern
  Recognition (CVPR)}, page 779–788. IEEE, Jun 2016.
\newblock ISBN 978-1-4673-8851-1.
\newblock \doi{10.1109/CVPR.2016.91}.
\newblock URL \url{http://ieeexplore.ieee.org/document/7780460/}.

\bibitem[Ren et~al.(2017)Ren, He, Girshick, and Sun]{ren2016faster}
S.~Ren, K.~He, R.~Girshick, and J.~Sun.
\newblock Faster r-cnn: Towards real-time object detection with region proposal
  networks.
\newblock \emph{IEEE Transactions on Pattern Analysis and Machine
  Intelligence}, 39\penalty0 (6):\penalty0 1137–1149, Jun 2017.
\newblock ISSN 0162-8828, 2160-9292.
\newblock \doi{10.1109/TPAMI.2016.2577031}.

\bibitem[Ronneberger et~al.(2015)Ronneberger, Fischer, and
  Brox]{DBLP:journals/corr/RonnebergerFB15}
O.~Ronneberger, P.~Fischer, and T.~Brox.
\newblock \emph{U-Net: Convolutional Networks for Biomedical Image
  Segmentation}, volume 9351 of \emph{Lecture Notes in Computer Science}, page
  234–241.
\newblock Springer International Publishing, 2015.
\newblock ISBN 978-3-319-24573-7.
\newblock \doi{10.1007/978-3-319-24574-4_28}.
\newblock URL \url{http://link.springer.com/10.1007/978-3-319-24574-4_28}.

\bibitem[Sabour et~al.(2017)Sabour, Frosst, and
  Hinton]{sabourDynamicRoutingCapsules2017}
S.~Sabour, N.~Frosst, and G.~E. Hinton.
\newblock Dynamic {{Routing Between Capsules}}.
\newblock In \emph{Proceedings of the 31st {{International Conference}} on
  {{Neural Information Processing Systems}}}, {{NIPS}}'17, pages 3859--3869,
  {USA}, 2017. {Curran Associates Inc.}
\newblock ISBN 978-1-5108-6096-4.

\bibitem[Schmidt et~al.(2018)Schmidt, Weigert, Broaddus, and
  Myers]{schmidt2018_CellDetection}
U.~Schmidt, M.~Weigert, C.~Broaddus, and G.~Myers.
\newblock \emph{Cell Detection with Star-Convex Polygons}, volume 11071 of
  \emph{Lecture Notes in Computer Science}, page 265–273.
\newblock Springer International Publishing, 2018.
\newblock ISBN 978-3-030-00933-5.
\newblock \doi{10.1007/978-3-030-00934-2_30}.
\newblock URL \url{http://link.springer.com/10.1007/978-3-030-00934-2_30}.

\bibitem[Stacke et~al.(2019)Stacke, Eilertsen, Unger, and
  Lundstr{\"{o}}m]{DBLP:journals/corr/abs-1909-11575}
K.~Stacke, G.~Eilertsen, J.~Unger, and C.~Lundstr{\"{o}}m.
\newblock A closer look at domain shift for deep learning in histopathology.
\newblock \emph{CoRR}, abs/1909.11575, 2019.
\newblock URL \url{http://arxiv.org/abs/1909.11575}.

\bibitem[Thierbach et~al.(2018)Thierbach, Bazin, Back, Gavriilidis, Kirilina,
  Jäger, Morawski, Geyer, Weiskopf, and Scherf]{thierbach2018_CombiningDeepa}
K.~Thierbach, P.-L. Bazin, W.~d. Back, F.~Gavriilidis, E.~Kirilina, C.~Jäger,
  M.~Morawski, S.~Geyer, N.~Weiskopf, and N.~Scherf.
\newblock \emph{Combining Deep Learning and Active Contours Opens The Way to
  Robust, Automated Analysis of Brain Cytoarchitectonics}, volume 11046 of
  \emph{Lecture Notes in Computer Science}, page 179–187.
\newblock Springer International Publishing, 2018.
\newblock ISBN 978-3-030-00918-2.
\newblock \doi{10.1007/978-3-030-00919-9_21}.
\newblock URL \url{http://link.springer.com/10.1007/978-3-030-00919-9_21}.

\bibitem[Xie et~al.(2020)Xie, Sun, Song, Wang, Liu, Liang, Shen, and
  Luo]{xie2020_PolarMaskSingle}
E.~Xie, P.~Sun, X.~Song, W.~Wang, X.~Liu, D.~Liang, C.~Shen, and P.~Luo.
\newblock Polarmask: Single shot instance segmentation with polar
  representation.
\newblock In \emph{2020 IEEE/CVF Conference on Computer Vision and Pattern
  Recognition (CVPR)}, page 12190–12199. IEEE, Jun 2020.
\newblock ISBN 978-1-72817-168-5.
\newblock \doi{10.1109/CVPR42600.2020.01221}.
\newblock URL \url{https://ieeexplore.ieee.org/document/9157078/}.

\bibitem[Yagi(2011)]{yagi_color_2011}
Y.~Yagi.
\newblock Color standardization and optimization in {Whole} {Slide} {Imaging}.
\newblock \emph{Diagnostic Pathology}, 6\penalty0 (S1):\penalty0 S15, Dec.
  2011.
\newblock ISSN 1746-1596.
\newblock \doi{10.1186/1746-1596-6-S1-S15}.
\newblock URL
  \url{https://diagnosticpathology.biomedcentral.com/articles/10.1186/1746-1596-6-S1-S15}.

\bibitem[Yang et~al.(2020)Yang, Ghosh, Franklin, Chen, You, Narayan, Melcher,
  and Liphardt]{yang2020_NuSeTDeep}
L.~Yang, R.~P. Ghosh, J.~M. Franklin, S.~Chen, C.~You, R.~R. Narayan, M.~L.
  Melcher, and J.~T. Liphardt.
\newblock {{NuSeT}}: {{A}} deep learning tool for reliably separating and
  analyzing crowded cells.
\newblock \emph{PLOS Computational Biology}, 16\penalty0 (9):\penalty0
  e1008193, Sept. 2020.
\newblock ISSN 1553-7358.
\newblock \doi{10.1371/journal.pcbi.1008193}.

\bibitem[Zabawa et~al.(2020)Zabawa, Kicherer, Klingbeil, Töpfer, Kuhlmann, and
  Roscher]{Zabawa_2020}
L.~Zabawa, A.~Kicherer, L.~Klingbeil, R.~Töpfer, H.~Kuhlmann, and R.~Roscher.
\newblock Counting of grapevine berries in images via semantic segmentation
  using convolutional neural networks.
\newblock \emph{ISPRS Journal of Photogrammetry and Remote Sensing},
  164:\penalty0 73–83, Jun 2020.
\newblock ISSN 0924-2716.
\newblock \doi{10.1016/j.isprsjprs.2020.04.002}.
\newblock URL \url{http://dx.doi.org/10.1016/j.isprsjprs.2020.04.002}.

\bibitem[Zhang et~al.(2018)Zhang, Bai, Liao, Urtasun, Marcos, Tuia, and
  Kellenberger]{zhang2018_LearningDeep}
L.~Zhang, M.~Bai, R.~Liao, R.~Urtasun, D.~Marcos, D.~Tuia, and B.~Kellenberger.
\newblock Learning deep structured active contours end-to-end.
\newblock In \emph{2018 IEEE/CVF Conference on Computer Vision and Pattern
  Recognition}, page 8877–8885. IEEE, Jun 2018.
\newblock ISBN 978-1-5386-6420-9.
\newblock \doi{10.1109/CVPR.2018.00925}.
\newblock URL \url{https://ieeexplore.ieee.org/document/8579023/}.

\end{thebibliography}
